\newif\iftecrep\tecreptrue         

\documentclass[envcountsame]{llncs}
\usepackage{latexsym,amsmath,amssymb}
\iftecrep\fi

\def\xwidehat{}
\def\FOE{{\xwidehat{\mbox{\textit{F\kern-0.12emo\kern-0.08emE}}}}}
\def\FPL{{\xwidehat{\mbox{\textit{F\kern-0.08emP\kern-0.08emL}}}}}
\def\IFPL{{\xwidehat{\mbox{\textit{IF\kern-0.08emP\kern-0.08emL}}}}}
\def\best{{\mbox{\scriptsize\textit{b\kern-0.08eme\kern-0.05ems\kern-0.05emt}}}}
\def\foe{^{\xwidehat{\mbox{\scriptsize\textit{F\kern-0.13emo\kern-0.13emE}}}}}
\def\foett{^{\xwidehat{\mbox{\scriptsize\textit{F\kern-0.13emo\kern-0.13emE}}}_{\tilde T}}}
\def\fpl{^{\xwidehat{\mbox{\scriptsize\textit{F\kern-0.15emP\kern-0.15emL}}}}}
\def\ifpl{^{\xwidehat{\mbox{\scriptsize\textit{I\kern-0.08emF\kern-0.15emP\kern-0.15emL}}}}}
\def\FOEtilde{{\widetilde{\mbox{\textit{F\kern-0.12emo\kern-0.08emE}}}}}
\def\foetilde{^{\widetilde{\mbox{\scriptsize\textit{F\kern-0.13emo\kern-0.13emE}}}}}
\def\sfoe{^{\xwidehat{\mbox{\tiny\textit{F\kern-0.12emo\kern-0.18emE}}}}}
\def\sfpl{^{\xwidehat{\mbox{\tiny\textit{F\kern-0.2emP\kern-0.2emL}}}}}
\def\sifpl{^{\xwidehat{\mbox{\tiny\textit{I\kern-0.1emF\kern-0.2emP\kern-0.2emL}}}}}

\def\e{{\rm e}}                        
\def\approxleq{\;\mbox{\raisebox{-0.8ex}{$\stackrel{\displaystyle<}\sim$}}\;}
\def\ii{\hspace*{1.5em}}
\def\wprob{\mbox{ w.p. }}

\newlength\algowidth
\def\absatz#1{\vspace{0.5ex}\par\noindent\textbf{#1}}

\def\loss{\ell}
\def\Loss{L}

\newenvironment{Lemma} {\begin{lemma}} {\end{lemma}}
\newenvironment{Prop} {\begin{proposition}}
{\end{proposition}}
\newenvironment{Theorem} {\begin{theorem}} {\end{theorem}}
\newenvironment{Cor} {\begin{corollary}} {\end{corollary}}

\newenvironment{Proof} {\begin{proof}} {\ \nolinebreak
\hfill $\Box$ \vspace{2ex} \end{proof}}

\def\baq#1\eaq{\begin{align}#1\end{align}}
\def\baqn#1\eaqn{\begin{align*}#1\end{align*}}
\def\beq#1\eeq{\begin{equation}#1\end{equation}}
\def\beqn#1\eeqn{\begin{displaymath}#1\end{displaymath}}
\def\bqa#1\eqa{\begin{eqnarray}#1\end{eqnarray}}
\def\bqan#1\eqan{\begin{eqnarray*}#1\end{eqnarray*}}

\def\calA{\mathcal A}
\def\calB{\mathcal B}

\def\NNN{\mathbb N}
\def\RRR{\mathbb R}

\def\Expect{{\mathbf E}}
\def\Prob{{\mathbf P}}

\def\eps{\varepsilon}

\def\leqt{_{1:t}}
\def\ltt{_{<t}}

\def\leqT{_{1:T}}
\def\ltT{_{<T}}
\def\_norm{_\mathrm{norm}}

\newcommand{\zwidths}[1]{\rlap{$\scriptstyle #1$}}

\def\for_all{\mbox{ for all }}
\def\such_that{\mbox{ such that }}

\def\und{\mbox{ and }}

\def\eins{1\hspace{-0.23em}{\rm I}}
\def\tfrac#1#2{{\textstyle\frac{#1}{#2}}}
\def\tsqrt#1{{\textstyle\sqrt{#1}}}

\begin{document}

\title{\iftecrep\Large\hrule height5pt \vskip 6mm\fi
Defensive Universal Learning with Experts
\iftecrep\vskip 6mm \hrule height2pt\fi
}
\titlerunning{Defensive Universal Learning with Experts}

\author{Jan Poland\inst{1} \and Marcus Hutter\inst{2}}
\tocauthor{Jan Poland (Hokkaido University)}
\authorrunning{Jan Poland and Marcus Hutter}

\institute{ Grad.\ School of Inf.\ Sci.\ and Tech.,
 Hokkaido University, Japan \\
 \email{jan@ist.hokudai.ac.jp},
\texttt{http://www-alg.ist.hokudai.ac.jp/\~{}jan}
\and
IDSIA, Galleria 2, CH-6928 Manno (TI),
Switzerland\\
\email{marcus@idsia.ch},
\texttt{http://www.idsia.ch/\~{}marcus}}

\maketitle

\iftecrep\centerline{\sc TCS-TR-A-05-4 \ \ \ --- \ \ \ July 2005 \ \ \ --- \ \ \ IDSIA-15-05}\fi

\begin{abstract}
This paper shows how universal learning can be achieved with
expert advice. To this aim, we specify an experts algorithm with
the following characteristics: (a) it uses only feedback from the
actions actually chosen (bandit setup), (b) it can be applied with
countably infinite expert classes, and (c) it copes with losses
that may grow in time appropriately slowly. We prove loss bounds
against an adaptive adversary. From this, we obtain a master
algorithm for ``reactive" experts problems, which means that the
master's actions may influence the behavior of the adversary. Our
algorithm can significantly outperform standard experts algorithms
on such problems. Finally, we combine it with a universal expert
class. The resulting universal learner performs -- in a certain
sense -- almost as well as any computable strategy, for any online
decision problem. We also specify the (worst-case) convergence
speed, which is very slow.
\iftecrep

{\bf Keywords.} Prediction with expert advice, responsive
environments, partial observation game, bandits, universal
learning, asymptotic optimality.
\fi
\end{abstract}

\section{Introduction}

Expert advice has become a well-established paradigm of machine
learning in the last decade, in particular for prediction. It is
very appealing from a theoretical point of view, as performance
guarantees usually hold in the worst case, without any
(statistical) assumption on the data. Such assumptions are
generally required for other statistical learning methods, often
however not resulting in stronger guarantees.

Using expert advice in the standard way seems a rather bad idea in
some cases where the decisions of the learner or master algorithm
influence the behavior of the environment or adversary. One
example is the repeated prisoner's dilemma when the opponent plays
``tit for tat" (see Section \ref{sec:active}). This was noted and
resolved by \cite{Farias:03}, who introduced a ``strategic expert
algorithm" for so-called reactive environments. Their algorithm
works with a finite class of experts and attains asymptotically
optimal behavior. No convergence speed is asserted, and the
analysis is quite different from that of standard experts
algorithms.

In this paper, we show how the more general task with a countably
infinite expert class can be accomplished, building on standard
experts algorithms, and simultaneously also bounding the
convergence rate ($t^{-\frac{1}{10}}$, which can be
actually improved to $t^{-\frac{1}{3}+\eps}$). To this aim,
we will combine techniques from
\cite{Hannan:57,Kalai:03,McMahan:04,Hutter:04expert}
and obtain a master algorithm which performs well on loss
functions that may \emph{increase in time}. Then this is
applied to (possibly) reactive problems by yielding the
control to the selected expert
for an increasing period of time steps. Using a universal expert
class defined by the countable set of all programs on some fixed
universal Turing machine, we obtain an algorithm which is in a
sense asymptotically optimal with respect to \emph{any} computable
strategy. An easy additional construction guarantees that our
algorithm is computable, in contrast to other universal approaches
which are non-computable \cite{Hutter:04uaibook}. To our
knowledge, we also propose the first algorithm for  non-stochastic
bandit problems with countably many arms.

The paper is structured as follows. Section \ref{sec:algorithm}
introduces the problem setup, the notation, and the algorithm. In
Sections \ref{sec:master}, we give the (worst-case) analysis of
the master algorithm. The implications to active experts problems
and a universal master algorithms are given in Section
\ref{sec:active}. We discuss our results in Section
\ref{sec:discussion}.

\section{The master algorithm}\label{sec:algorithm}

\absatz{Setup}. We are acting in an online decision problem. ``We"
is here an abbreviation for the master algorithm which is to be
designed. An ``online decision problem" is to be understood in a
very general sense, it is just a sequence of decisions each of
which results in some loss. This could be e.g.\ a prediction task,
a repeated game, etc. In each round, that is at each time step
$t$, we have access to the recommendations of countably infinitely
many ``experts" or strategies. (For simplicity, we restrict our
notation to a countably infinite expert class, all results also
hold for finite classes.) We do not specify what exactly a
``recommendation" is -- we just follow the advice of one expert.
\emph{Before} we reveal our move, the adversary has to assign
losses $\loss_t^i\geq 0$ to \emph{all} experts $i$. There is an
upper bound $B_t\geq\|\loss_t\|_\infty$ on the maximum loss the
adversary may use. This quantity may depend on $t$ and is
not controlled by the adversary. After the move, only the
loss of the
selected expert $i$ is revealed. Our goal is to perform nearly as
well as the best available strategy (expert) in terms of
cumulative loss, after any number $T$ of time steps which is not
known in advance. The difference between our loss and the loss of
some expert is also termed \emph{regret}. We consider the general
case of an \emph{adaptive} adversary, which may assign losses
depending on our past decisions.

If there are only finitely many experts or strategies, then it is
common to give no prior preferences to any of them. Formally, this
is realized by defining uniform \emph{prior weights}
$w^i=\frac{1}{n}$ for each expert $i$. This is not possible for
countably infinite expert classes, as there is no uniform
distribution on the natural numbers. In this case, we need some
non-uniform prior $(w^i)_{i\in\NNN}$ and require $w^i>0$
for all
experts $i$ and $\sum_i w^i\leq 1$. We also define the complexity
of expert $i$ as $k^i=-\ln w^i$. This quantity is important since
in the full observation game (i.e.\ after our decision we get to
know the losses of \emph{all} experts), the regret can usually be
bounded by some function of the best expert's complexity.

Our algorithm ``Follow or Explore" ($\FOE$, specified in Fig.
\ref{fig:foe}) builds on McMahan and Blum's online geometric
optimization algorithm \cite{McMahan:04}. It is a bandit version
of a ``Follow the Perturbed Leader" experts algorithm. This
approach to online prediction and playing repeated games has been
pioneered by Hannan \cite{Hannan:57}. For the full observation
game and uniform prior, \cite{Kalai:03} gave a very elegant
analysis which is clearly different from the standard analysis of
exponential weighting schemes. It has one advantage over other
aggregating algorithms such as exponential weighting schemes: the
analysis is not complicated if the learning rate is dynamic rather
than fixed in advance. A dynamic learning rate is necessary if
there is no target time $T$ known in advance. For non-uniform
prior, an analysis was given in \cite{Hutter:04expert}. The
following issues are important for $\FOE$'s design.

\begin{figure}[t!]
\begin{center}
\fbox{
\begin{minipage}{\algowidth}
For $t=1,2,3,\ldots$\\
\ii set $\hat\loss_t^i=\hat B_t$ for $i\not\in\{t\geq\tau\}$
 (see (\ref{eq:maxest}))\\
\ii sample $r_t\in\{0,1\}$ independently s.t.
$P[r_t=1]=\gamma_t$\\
\ii If $r_t=0$ Then\\
\ii \ii invoke $\FPL(t)$ and play its decision\\
\ii \ii set $\hat\loss_t^i=0$ for $i\in\{t\geq\tau\}$\\
\ii Else\\
\ii \ii sample $I_t\foe:=u_t$ ``uniformly", see
(\ref{eq:unonu}),
and play $I:=I_t\foe$\\
\ii \ii set $\hat\loss_t^I=\loss_t^I/(u_t^I \gamma_t)$ and
$\hat\loss_t^i=0$ for $i\in\{t\geq\tau\}\setminus\{I\}$
\end{minipage}}
\caption{The algorithm $\FOE$. The exploration rate
$\gamma_t$ will be
specified in Corollary \ref{cor:arbitrary}.}
\label{fig:foe}
\end{center}
\end{figure}

\begin{figure}[t!]
\begin{center}
\fbox{
\begin{minipage}{\algowidth}
Sample $q_t^i\stackrel{d.}{\sim}\mbox{\textit{Exp}}$
(i.e.\ $\Prob(q_t^i\geq x)=e^{-x}$ for $x\geq 0$) indep.\
$\forall i\in\{t\geq\tau\}$\\
select and play
$I_t\fpl=\arg\min\limits_{i:t\geq\tau}
\{\eta_t\hat \loss^i\ltt+k^i-q_t^i\}$
\end{minipage}}
\caption{The algorithm $\FPL$. The learning rate $\eta_t$ will be
specified in Corollary \ref{cor:arbitrary}.}
\label{fig:fpl}
\end{center}
\end{figure}

\absatz{Exploration}. Since we are playing the bandit game (as
opposed to the full information game), we need to explore
sufficiently \cite{Auer:95,McMahan:04}. At each time step $t$, we
decide randomly according to some exploration rate
$\gamma_t\in(0,1)$ whether to explore or not. If so, we would like
to choose an expert according to the prior distribution. There is
a caveat: In order to make the analysis go through, we have to
assure that we are working with \emph{unbiased} estimates of the
losses. This is achieved by dividing the observed loss by the
probability of choosing the expert. But this quantity could become
arbitrarily large if we admit arbitrarily small weights. We
address this problem by \emph{finitizing} the expert pool at each
time $t$. For each expert $i$, we define an \emph{entering time}
$\tau^i$, that is, expert $i$ is active only for $t\geq\tau^i$. We
denote the set of active experts at time $t$ by
$\{t\geq\tau\}=\{i:t\geq\tau^i\}$. For exploration, the prior is
then replaced by the finitized prior distribution $u_t$,
\beq
\label{eq:unonu}
\Prob(u_t=i)={w^i\eins_{t\geq\tau^i}\over\sum_j
w^j\eins_{t\geq\tau^j}}.
\eeq
Consequently, the maximum unbiasedly estimated instantaneous loss
is (note that the exploration probability also scales with the
exploration rate $\gamma_t$)
\beq
\label{eq:maxest}
\hat B_t=\frac{B_t}{\gamma_t\min\{w^i:t\geq\tau^i\}}.
\eeq
It is convenient for the analysis to assign estimated loss of
$\hat B_t$ to all currently inactive experts. Observe finally that
in this way, our master algorithm $\FOE$ always deals with a
finite expert class and is thus computable.

\absatz{Follow the perturbed leader} ($\FPL$, specified in Fig.
\ref{fig:fpl}) is invoked if $\FOE$ does not explore. Just
following the ``leader" (the best expert so far) may not be a good
strategy \cite{Kalai:03}. Instead we subtract an exponentially
distributed perturbation $q_t$ from the current score (the
complexity penalized past loss) of the experts. An important
detail of the $\FPL$ subroutine is the \emph{learning rate}
$\eta_t>0$, which should be adaptive if the total number of steps
$T$ is not known in advance. Please see e.g.\
\cite{Kalai:03,Hutter:04expert} for more details. Also the variant
of $\FPL$ we use (specified in Fig. \ref{fig:fpl}) works on the
finitized expert pool.

Note that each time randomness is used, it is assumed to be
\emph{independent} of the past randomness. Performance is
evaluated in terms of true or estimated cumulative loss, this is
specified in the notation. E.g.\ for the true loss of $\FPL$ up to
and including time $T$ we write $\loss\fpl\leqT$, while the
estimated loss of $\FOE$ and not including time $T$ is $\hat
\loss\foe\ltT$.

\section{Analysis on the master's time scale}\label{sec:master}

The following analysis uses McMahan and Blum's trick
\cite{McMahan:04} in order to prove bounds against adaptive
adversary. With a different argument, it is possible to
circumvent Lemma \ref{lemma:behbeh}, thus achieving better
bounds \cite{Poland:05fpla}. This will be briefly discussed
in the last section.

Let $B_t\geq 0$ be some sequence of upper bounds on the
instantaneous losses, $\gamma_t\in(0,1)$ be a sequence of
exploration rates, and $\eta_t>0$ be a \emph{decreasing}
sequence of learning rates.
The analysis proceeds according to the following diagram
(where $\Loss$ is an informal abbreviation for the loss and
always refers to cumulative loss, but sometimes additionally
to instantaneous loss).
\baq
\label{eq:diagram}
\Loss\foe \approxleq \Expect \Loss\foe \approxleq \Expect \Loss\fpl
\approxleq \Expect \hat \Loss\fpl
\approxleq \Expect \hat \Loss\ifpl \approxleq \Expect \hat \Loss^{\xwidehat\best}
\approxleq \Loss^{\best}
\eaq
Each ``$\approxleq$" means that we bound the quantity on the left
by the quantity on the right plus some additive term. The first
and the last expressions are the losses of the $\FOE$ algorithm
and the best expert, respectively. The intermediate quantities
belong to different algorithms, namely $\FOE$, $\FPL$, and a third
one called $\IFPL$ for ``infeasible" FPL \cite{Kalai:03}. $\IFPL$,
as specified in Fig. \ref{fig:ifpl}, is the same as $\FPL$ except
that it has access to an oracle providing the current estimated
loss vector $\hat\loss_t$ (hence infeasible). Then it assigns
scores of $\eta_t\hat \loss^i\leqt+k^i-q_t^i$ instead of
$\eta_t\hat \loss^i\ltt+k^i-q_t^i$.

\begin{figure}[t!]
\begin{center}
\fbox{
\begin{minipage}{\algowidth}
Sample $q_t^i\stackrel{d.}{\sim}\mbox{\textit{Exp}}$
independently for all $i\in\{t\geq\tau\}$\\
select and play
$I_t\fpl=\arg\min\limits_{i:t\geq\tau}
\{\eta_t\hat \loss^i\leqt+k^i-q_t^i\}$\\
\end{minipage}}
\caption{The algorithm $\IFPL$. The learning rate $\eta_t$
will be specified in Corollary \ref{cor:arbitrary}.}
\label{fig:ifpl}
\end{center}
\end{figure}

The randomization of $\FOE$ and $\FPL$ gives rise to two
filtrations of $\sigma$-algebras. By $\calA_t$ we denote the
$\sigma$-algebra generated by the $\FOE$'s randomness up to
time $t$, meaning
\emph{only} the random variables $\{u_{1:t},r_{1:t}\}$. Then
$(\calA_t)_{t\geq 0}$ is a filtration ($\calA_0$ is the
trivial
$\sigma$-algebra). We may also write $\calA=\bigcup_{t\geq
0}\calA_t$. Similarly, $\calB_t$ is the $\sigma$-algebra generated
by the $\FOE$'s \emph{and} $\FPL$'s randomness up to time $t$
(i.e.\ $\calB_t\widehat=\{u_{1:t},r_{1:t},q_{1:t}\}$).
Then clearly $\calA_t\subset\calB_t$ for each $t$.

The reader should think of the expectations in
(\ref{eq:diagram}) as of both ordinary and
\emph{conditional} expectations. Conditional expectations
are mostly with respect to $\FOE$'s past randomness
$\calA_{t-1}$. These conditional expectations of some random
variable $X$ are abbreviated by
\baqn
  \Expect_t[X]:=\Expect[X|\calA_{t-1}].
\eaqn
Then $\Expect_t[X]$ is an
$\calA_{t-1}$-measurable random variable, meaning that its value
is determined for fixed past randomness $\calA_{t-1}$. Note in
particular that the estimated loss vectors $\hat \loss_t^i$ are
random vectors which depend on $\FOE$'s randomness $\calA_t$ up to
time $t$. In this way, $\FOE$'s (and $\FPL$'s and
$\IFPL$'s) actions depend on $\FOE$'s past randomness. Note,
however, that they do not depend on $\FPL$'s randomness $q_{1:t}$.

We now start with proving the diagram (\ref{eq:diagram}). In
order to understand the analysis, it is important to consider
each intermediate algorithm as a stand-alone procedure which
is actually executed (with an oracle if necessary) on the
specified inputs (e.g.\ on the estimated losses) and has the
asserted performance guarantees (e.g.\ again on the
estimated losses).

\begin{Lemma} \label{lemma:foefoe}
$\big[\Loss\foe \approxleq \Expect \Loss\foe\big]$ For each $T\geq
1$ and $\delta_T\in(0,1)$, with probability at least
$1-\tfrac{\delta_T}{2}$ we have \baqn
\loss\foe\leqT\leq\sum_{t=1}^t\Expect_t\loss\foe_t
+\tsqrt{(2\ln\tfrac{4}{\delta_T}){\textstyle\sum\nolimits_{t=1}^{T}
B_t^2}}. \eaqn
\end{Lemma}

\begin{Proof} The sequence of random variables
$X_T=\sum_{t=1}^T\big[\loss\foe_t-
\Expect_t\loss\foe_t\big]$ is a martingale with
respect to the filtration $\calB_t$ (not $\calA_t$!). In
order to see this, observe
$\Expect[\loss\foe_T|\calB_{T-1}]=
\Expect\big(\Expect[\loss\foe_T|\calA_{T-1}]\big|\calB_{T-1}\big)$
and
$\Expect[\loss\foe_t|\calB_{T-1}]=\loss_t\foe$
for $t<T$, which implies
\bqan
\Expect[X_T|\calB_{T-1}] &=&
\sum\nolimits_{t=1}^T\left(
\Expect[\loss\foe_t|\calB_{T-1}] -
\Expect\big[\Expect[\loss\foe_t|\calA_{t-1}]\big|\calB_{T-1}\big]\right)
\\ &=&
\sum_{t=1}^{T-1}\nolimits\left(
\loss\foe_t-
\Expect[\loss\foe_t|\calA_{t-1}]\right)=X_{T-1}.
\eqan
Its differences are bounded: $|X_t-X_{t-1}|\leq B_t$. Hence,
it follows from Azuma's inequality (see e.g.\
\cite{Motwani:95}) that the probability that $X_T$ exceeds
some $\lambda>0$ is bounded by
$p=2\exp\big(-\tfrac{\lambda^2}{2\sum_t B_t^2}\big)$.
Requesting $\tfrac{\delta_T}{2}=p$ and solving for $\lambda$
gives the assertion.
\end{Proof}

\begin{Lemma} \label{lemma:foefpl}
$\big[\Expect \loss\foe \approxleq \Expect \loss\fpl\big]$
$\Expect_t
\loss_t\foe\leq(1-\gamma_t)\Expect_t\loss_t\fpl+\gamma_tB_t$
holds $\forall t\geq 1$.
\end{Lemma}

This follows immediately from the specification of $\FOE$.
Clearly, a corresponding assertion for the ordinary expectations
holds by just taking expectations on both sides. This is the case
for all subsequent lemmas, except for Lemma \ref{lemma:behbeh}.

The next lemma relating $\Expect \Loss\fpl$ and
$\Expect \hat \Loss\fpl$ is technical but intuitively clear. It
states that in expectation, the real loss suffered by
$\FPL$ is the same as the estimated loss. This is
simply because the loss estimate is unbiased. A combination of
this and the previous lemma was shown in \cite{McMahan:04}.
Note that $\hat\loss_t\fpl$ is the loss $\hat\loss_t^I$ estimated by
$\FOE$, but for the expert $I=I_t\fpl$ chosen by $\FPL$.

\begin{Lemma} \label{lemma:fplfpl}
$\big[\Expect \Loss\fpl \approxleq \Expect \hat \Loss\fpl\big]$ For
each $t\geq 1$, we have
$\Expect_t \loss_t\fpl=\Expect_t\hat\loss_t\fpl$.
\end{Lemma}

\begin{Proof} Let $f^i_t=\Prob[I_t\fpl=i|\calA_{t-1}]$
be the probability distribution over actions $i$ which $\FPL$ uses
at time $t$, depending on the past randomness $\calA_{t-1}$. Let
$u_t$ be the finitized prior distribution (\ref{eq:unonu}) at time
$t$. Then
\beqn
\Expect_t[\hat\loss_t\fpl]
(1-\gamma_t)\cdot 0 +
\gamma_t \sum_{i=1}^\infty  f_t^i [(1-u_t^i)\cdot 0+u_t^i \hat\loss_t^i|_{r_t=1\wedge I\sfoe_t=i}]
\sum_{i=1}^\infty  f_t^i\loss_t^i =
\Expect_t[\loss_t\fpl],
\eeqn
where $\hat\loss_t^i|_{r_t=1\wedge
I\sfoe_t=i}=\loss_t^i/(u_t^i\gamma_t)$ is the estimated loss under
the condition that $\FOE$ decided to explore ($r_t=1$) and chose
action $I\foe_t=i$.
\end{Proof}

The following lemma relates the losses of $\FPL$ and
$\IFPL$. It is proven in \cite{Kalai:03} and
\cite{Hutter:04expert}. We give the full proof,
since it is the only step in the analysis where we
have to be careful with the upper loss bound $B_t$. Let
$\hat B_t$ be the upper bound on the estimated loss
(\ref{eq:maxest}). (We remark that also for
\emph{weighted averaging forecasters}, losses which grow
sufficiently slowly do not cause any problem in the
analysis. In this way, it is straightforward to modify the
algorithm by Auer et al. \cite{Auer:02bandit} for reactive
tasks with a finite expert class.)

\begin{Lemma} \label{lemma:fplifpl}
$\big[\Expect \hat \Loss\fpl \approxleq \Expect \hat \Loss\ifpl\big]$
For all $t\geq 1$,
$\Expect_t \hat\loss_t\fpl\leq\Expect_t\hat\loss_t\ifpl+
\gamma_t\eta_t \hat B_t^2$ holds.
\end{Lemma}

\begin{Proof}
If $r_t=0$, $\hat\loss_t= 0$ and thus
$\hat\loss_t\fpl=\hat\loss_t\ifpl$ holds. This happens with
probability $1-\gamma_t$. Otherwise we have
\beq
\label{eq:efpl}
\Expect_t\hat\loss_t\fpl=
\sum_{i=1}^\infty\int\eins_{I\sfpl_t=i}
\hat\loss_t^i d\mu(x),
\eeq
where $\mu$ denotes the (exponential) distribution of the
perturbations, i.e.\ $x_i:=q_t^i$ and density
$\mu(x):=\e^{-\|x\|_\infty}$. The idea is now that if action $i$
was selected by $\FPL$, it is -- because of the exponentially
distributed perturbation -- with high probability also selected by
$\IFPL$. Formally, we write $u^+=\max(u,0)$ for $u\in\RRR$,
abbreviate $\lambda=\hat \loss\ltt+k/\eta_t$, and denote by
$\int\ldots d\mu(x_{\neq i})$ the integration leaving out the
$i$th action. Then, using
$\eta_t\lambda_i-x_i\leq\eta_t\lambda_j-x_j$ $\forall j$ if
$I_t\fpl=i$ in the first equation, and
$\hat B_t\geq \hat\loss_t^i-\hat\loss_t^j$ in the last line, we
get
\baqn
\label{eq:1}
\int\eins_{I\sfpl_t=i}\hat\loss_t^i d\mu(x)
 &=
  \int\!\int\limits_{x_i\geq\zwidths{\max\limits_{j\neq
  i}\{\eta_t(\lambda_i-\lambda_j)+x_j\}}}
  \hat\loss_t^i d\mu(x_i) d\mu(x_{\neq i})=
\int\hat\loss_t^i\;\e^{-(\max\limits_{j\neq
i}\{\eta_t(\lambda_i-\lambda_j)+x_j\})^+}\!
   d\mu(x_{\neq i})\\
& \leq  \int\hat\loss_t^i\;\e^{\eta_t \hat B_t}\e^{-(\max\limits_{j\neq i}\{\eta_t(\lambda_i-\lambda_j)+x_j\}+\eta_t \hat B_t)^+}
   d\mu(x_{\neq i})\\
& \leq\e^{\eta_t \hat B_t} \int\hat\loss_t^i\;\e^{-(\max\limits_{j\neq i}\{\eta_t(\lambda_i+\hat\loss_t^i-\lambda_j-\hat\loss_t^j)+x_j\})^+}
   d\mu(x_{\neq i})\\
& =\e^{\eta_t \hat B_t}\int\eins_{I\sifpl_t=i}
\hat\loss_t^i d\mu(x).
\eaqn
Summing over $i$ and using the analog of (\ref{eq:efpl})
for $\IFPL$, we see that if $r_t=1$, then
$\Expect_t\hat\loss_t\fpl\leq
\e^{\eta_t \hat B_t}\Expect_t\hat\loss_t\ifpl$ holds. Thus
$\Expect_t\hat\loss_t\ifpl\geq
\e^{-\eta_t \hat B_t}\Expect_t\hat\loss_t\fpl
\geq(1-\eta_t \hat B_t)\Expect_t\hat\loss_t\fpl
\geq\Expect_t\hat\loss_t\fpl-\eta_t\hat B_t^2$.
The assertion now follows by taking expectations w.r.t.\ $r_t$.
\end{Proof}

The next lemma relates the losses of $\IFPL$ and the best action
in hindsight. For an oblivious adversary (which means that the
adversary's decisions do not depend on our past actions), the
proof is quite simple \cite{Kalai:03}. An additional step is
necessary for an adaptive adversary \cite{Hutter:05expertx}.

\begin{Lemma} \label{lemma:ifplbeh}
$\big[\Expect \hat \Loss\ifpl \approxleq \Expect \hat \Loss^{\xwidehat\best}\big]$
Assume that $\sum_i\e^{-k^i}\leq 1$ and $\tau^i$ depends
monotonically on $k^i$, i.e.\
$\tau^i\geq\tau^j$ if and only if $k^i\geq k^j$. Assume
decreasing learning rate $\eta_t$.
For all $T\geq 1$ and all $i\geq 1$,
\beqn
\sum_{t=1}^T\Expect_t \hat \loss\ifpl_t\leq
\hat\loss\leqT^i+{\textstyle\frac{k^i+1}{\eta_T}}.
\eeqn
\end{Lemma}

\begin{Proof}
This is a modification of the corresponding proofs in
\cite{Kalai:03} and \cite{Hutter:04expert}. We may fix the
randomization $\calA$ and suppress it in the notation. Then
we only need to show
\beq
\label{eq:showtau}
\Expect \hat
\loss\ifpl\leqT\leq\min\limits_{i\geq 1}
\{\hat\loss\leqT^i+{\textstyle\frac{k^i+1}{\eta_T}}\},
\eeq
where the expectation is with respect to $\IFPL$'s randomness $q_{1:T}$.

Assume first that the adversary is oblivious. We define an
algorithm $A$ as a variant of $\IFPL$ which samples only one
perturbation vector $q$ in the beginning and uses this in each
time step, i.e.\ $q_t\equiv q$. Since the adversary is oblivious,
$A$ is equivalent to $\IFPL$ in terms of expected performance.
This is all we need to show (\ref{eq:showtau}). Let
$\eta_0=\infty$ and $\lambda_t=\hat\loss_t+(k-q)
\big(\frac{1}{\eta_t}-\frac{1}{\eta_{t-1}}\big)$, then
$\lambda\leqt=\hat\loss\leqt+\frac{k-q}{\eta_t}$. Recall
$\{t\geq\tau\}=\{i:t\geq\tau^i\}$. We argue by induction that for
all $T\geq 1$,
\beq
\label{eq:ifplbehtau}
\sum_{t=1}^T \lambda_t^A\leq
\min_{T\geq\tau}\lambda\leqT^i+
\max_{T\geq\tau}\big\{{\tfrac{q^i-k^i}{\eta_T}}\big\}.
\eeq
This clearly holds for $T=0$. For the induction step, we have
to show
\baq
\label{eq:showtau2}
\min_{T\geq\tau}\lambda\leqT^i+
\max_{T\geq\tau}\big\{{\tfrac{q^i-k^i}{\eta_T}}\big\}+
\lambda_{T+1}^A
&\leq
\lambda\leqT^{I^A_{T+1}}+
\max_{T+1\geq\tau}\big\{{\tfrac{q^i-k^i}{\eta_{T+1}}}\big\}+
\lambda_{T+1}^{I^A_{T+1}}\\
\nonumber
&=
\min_{T+1\geq\tau}\lambda_{1:T+1}^i+
\max_{T+1\geq\tau}\big\{{\tfrac{q^i-k^i}{\eta_{T+1}}}\big\}.
\eaq
The inequality is obvious if $I_{T+1}^A\in\{T\geq\tau\}$.
Otherwise, let
$J=\arg\max\big\{q^i-k^i:i\in\{T\geq\tau\}\big\}$. Then
\bqan
\min_{T\geq\tau}\lambda\leqT^i+
  \max_{T\geq\tau}\big\{{\tfrac{q^i-k^i}{\eta_T}}\big\}
\leq\textstyle
  \lambda\leqT^J+\tfrac{q^J-k^J}{\eta_T}
  & = & \sum_{t=1}^T\hat\loss_t^J
  \leq \sum_{t=1}^T\hat B_t\\
& = &
  {\textstyle\sum_{t=1}^T}\hat\loss_t^{I_{T+1}^A}
  \leq \lambda\leqT^{I^A_{T+1}}+
  \max_{T+1\geq\tau}\big\{{\tfrac{q^i-k^i}{\eta_{T+1}}}\big\}
\eqan
shows (\ref{eq:showtau2}). Rearranging terms in
(\ref{eq:ifplbehtau}), we see
\beqn
  \sum_{t=1}^T \hat\loss_t^A\leq
  \min_{T\geq\tau}\lambda\leqT^i \!+\!
  \max_{T\geq\tau^i}\big\{{\textstyle\frac{q^i-k^i}{\eta_T}}\big\}
  \!+\!\! \sum_{t=1}^T
(q\!-\!k)^{I_t^A}\big({\textstyle\frac{1}{\eta_t}\!-\!\frac{1}{\eta_{t-1}}}\big)
\eeqn
The assertion (\ref{eq:showtau}) -- still for oblivious
adversary and $q_t\equiv q$ -- then follows by taking
expectations and using
\bqa
\label{eq:showtau3}
\Expect\min_{T\geq\tau}\lambda\leqT^i
\ \leq  \ \min_{T\geq\tau}\{\hat\loss\leqT^i+\tfrac{k^i}{\eta_T}
-\Expect {\textstyle\frac{q^i}{\eta_T}} \}
&\leq& \min_{i\geq 1}\{\hat\loss\leqT^i+{\textstyle\frac{k^i-1}{\eta_T}}\} \und\\
\label{eq:showtau4}
\Expect\sum_{t=1}^T
(q-k)^{I_t^A}\big(\tfrac{1}{\eta_t}-\tfrac{1}{\eta_{t-1}}\big)
&\leq&
\Expect\max_{T\geq\tau}\big\{{\textstyle\frac{q^i-k^i}{\eta_T}}\big\}
\ \leq\ {\textstyle\frac{1}{\eta_T}}.
\eqa
The second inequality of (\ref{eq:showtau3}) holds because
$\tau^i$ depends monotonically on $k^i$, and $\Expect q^i=1$,
and maximality of $\hat\loss_{1:T}^i$ for $T<\tau_i$.
The second inequality of (\ref{eq:showtau4}) can be proven by
a simple application of the union bound, see e.g.\
\cite[Lem.1]{Hutter:04expert}.

Sampling the perturbations $q_t$ independently is equivalent under
expectation to sampling $q$ only once. So assume that $q_t$ are
sampled independently, i.e.\ that $\IFPL$ is played against an
oblivious adversary: (\ref{eq:showtau}) remains valid. In the last
step, we argue that then (\ref{eq:showtau}) also holds for an
\emph{adaptive} adversary. This is true because the future actions
of $\IFPL$ do not depend on its past actions, and therefore the
adversary cannot gain from deciding after having seen $\IFPL$'s
decisions. This argument can be made formal, as shown in
\cite[Lemma 12]{Hutter:05expertx}. (Note the subtlety that the
future actions of $\FOE$ would depend on its past actions.)
\end{Proof}

Finally, we give a relation between the estimated and true
losses (adapted from \cite{McMahan:04}).

\begin{Lemma} \label{lemma:behbeh}
$\big[\Expect \hat \Loss^{\xwidehat\best} \approxleq \Loss^{\best}\big]$
For each $T\geq 1$, $\delta_T\in(0,1)$, and $i\geq 1$,
we have
\bqan
(i) && \hat \loss\leqT^i
\leq \loss\leqT^i+
\sqrt{(2\ln\tfrac{4}{\delta_T})\textstyle\sum\nolimits_{t=1}^{T}
\hat B_t^2} +\textstyle\sum\nolimits_{t=1}^{\tau^i-1}\hat B_t
\mbox{ w.p. $1-\tfrac{\delta_T}{2}$ and hence}\\
(ii) && \Expect \hat \loss\leqT^i
\leq \loss\leqT^i+
\sqrt{(2\ln\tfrac{4}{\delta_T})\textstyle\sum\nolimits_{t=1}^{T}\hat B_t^2}+
\tfrac{\delta_T}{2} {\textstyle\textstyle\sum\nolimits_{t=1}^{T}\hat B_t}
+\textstyle\sum\nolimits_{t=1}^{\tau^i-1}\hat B_t.
\eqan
\end{Lemma}

\begin{Proof}
For $t\geq\tau^i$, $X_t=\hat \loss\leqt^i-\loss\leqt^i$ is a martingale, since
\beqn
\Expect[X_t|\calA_{t-1}]=\Expect[\hat \loss^i\leqt|\calA_{t-1}]-\loss\leqt^i
= X_{t-1}+\Expect[\hat\loss^i_t|\calA_{t-1}]-\loss_t^i=X_{t-1}.
\eeqn
It is clear that $X_{\tau^i-1}\leq\sum_{t=1}^{\tau^i-1}\hat B_t$.
Moreover, $|X_t-X_{t-1}|\leq\hat B_t$ for $t\geq\tau^i$, i.e.\ we
have bounded differences. By Azuma's inequality, the actual value
$X_T-X_{\tau^i-1}$ does not exceed
$\tsqrt{(2\ln\tfrac{4}{\delta_T}){\textstyle\sum\nolimits_{t=1}^{T}
\hat B_t^2}}$ with probability $1-\tfrac{\delta_T}{2}$. This
proves $(i)$. To arrive at $(ii)$, take expectations and observe
that $(i)$ fails with probability at most $\tfrac{\delta_T}{2}$,
in which case $\hat
\loss^i\leqT\leq{\textstyle\sum\nolimits_{t=1}^{T}\hat B_t}$
holds.
\end{Proof}

We now combine the above results and derive an upper bound
on the expected regret of $\FOE$ against an adaptive
adversary.

\begin{Theorem} \label{th:general} {\upshape[$\FOE$ against
an adaptive adversary]} Let $\sum_i\e^{-k^i}\leq 1$, $\tau^i$
depend monotonically on $k^i$, and the learning rate $\eta_t$ be
decreasing. Let $\loss_t$ be some possibly adaptive assignment of
(true) loss vectors satisfying $\|\loss_t\|_\infty\leq B_t$. Then
for all experts $i$, we have with probability at least
$1-\delta_T$
\beqn
\loss\foe\leqT \leq
\loss\leqT^i+\tfrac{k^i+1}{\eta_T}+
\sum_{t=1}^{\tau^i-1}\!\!\!\hat B_t +
\sum_{t=1}^T\!\!\!\gamma_t\eta_t\hat B_t^2
+\sum_{t=1}^T\!\!\!\gamma_t B_t +
\sqrt{(2\ln\tfrac{4}{\delta_T})}\!\left(\!
\sqrt{{\sum_{t=1}^{ T }\!\!\hat B_t}}
+\sqrt{{\sum_{t=1}^{T}\!\! B_t^2}}\right)\!.
\eeqn
Consequently, in expectation, we have
\beqn
\Expect \loss\foe\leqT \leq
\loss\leqT^i+\tfrac{k^i+1}{\eta_T}+
\sum_{t=1}^{\tau^i-1}\hat B_t +
\sum_{t=1}^T\gamma_t\eta_t\hat B_t^2
+\sum_{t=1}^T\gamma_t B_t +
\sqrt{(2\ln\tfrac{4}{\delta_T}){\sum_{t=1}^{T}
\hat B_t}}+
\tfrac{\delta_T}{2}
{\sum_{t=1}^{T}\hat B_t}.
\eeqn
\end{Theorem}

\begin{Proof}
This follows by summing up all excess terms in the above lemmas.
Recall that we only need to take expectations on both sides of the
assertions of Lemmas \ref{lemma:foefpl}--\ref{lemma:ifplbeh} in
order to obtain the second bound on the expectation (and we don't
need Lemma \ref{lemma:foefoe} there).
\end{Proof}

\begin{Cor} \label{cor:arbitrary}
Assume the conditions of Theorem \ref{th:general} and choose
$\eta_t=t^{-\frac{3}{4}}$ and $\gamma_t=t^{-\frac{1}{4}}$. Then
\bqan
(i) & B_t\equiv 1,\tau^i\!=\!\lceil(w^i)^{-8}\rceil
&\Rightarrow
\Expect \loss\foe\leqT\leq
\loss\leqT^i+O\big((\tfrac{1}{w^i})^{11} +
k^i T^{\frac{7}{8}}\sqrt{\ln T}\big),\\
(ii) & B_t\equiv 1,\tau^i\!=\!\lceil(w^i)^{-8}\rceil
&\Rightarrow
\loss\foe\leqT\leq
\loss\leqT^i+O\big((\tfrac{1}{w^i})^{11}\!\!+\!
k^i T^{\frac{7}{8}}\sqrt{\ln T}\big)\!\wprob\!
1\!-\!T^{-2},\\
(iii) &
B_t\!=\!t^{\frac{1}{16}},\tau^i\!=\!\lceil(w^i)^{-16}\rceil
&\Rightarrow
\Expect \loss\foe\leqT\leq
\loss\leqT^i+O\big((\tfrac{1}{w^i})^{22} +
k^i T^{\frac{7}{8}}\sqrt{\ln T}\big), \und\\
(iv) &
B_t\!=\!t^{\frac{1}{16}},\tau^i\!=\!\lceil(w^i)^{-16}\rceil
&\Rightarrow \loss\foe\leqT\leq
\loss\leqT^i+O\big((\tfrac{1}{w^i})^{22}\!\!+\!
k^i T^{\frac{7}{8}}\sqrt{\ln T}\big)\!\wprob\!1\!-\!T^{-2},
\eqan
for all $i$ and $T\geq 1$ (recall $k^i=-\ln w^i$).
Moreover, in both cases (bounded and growing
$B_t$) $\FOE$ is asymptotically optimal w.r.t.\
each expert, i.e.\ for all $i$,
\beqn
\limsup_{T\to\infty} \frac{\loss\foe\leqT-\loss^i\leqT}{T}\leq 0
\quad\mbox{ almost surely.}
\eeqn
\end{Cor}

The asymptotic optimality is sometimes termed
\emph{Hannan-consistency}, in particular if the limit equals
zero. We only show the upper bound.

\begin{Proof}
Assertions $(i)$-$(iv)$ follow from the previous theorem: Set
$\delta_T=T^{-2}$, abbreviate
$w^{\min}_T=\min\{w^i:t\geq\tau^i\}$, and observe that for
$\tau^i=\lceil(w^i)^{-\alpha}\rceil$ and $B_t=t^\beta$, we have
\baqn w^{\min}_T=\min\{w^i:T\geq\lceil(w^i)^{-\alpha}\rceil\}&\geq
\min\{w^i:T^{-\frac{1}{\alpha}}\leq w^i\rceil\}\geq
T^{-\frac{1}{\alpha}} \quad\und\\
\sum_{t=1}^{\tau^i-1}\hat B_t \leq
(\tau^i-1)\hat B_{\tau^i-1} &\leq
\tfrac{(w^i)^{-\alpha} B_{\tau^i-1}}{\gamma_{\tau^i-1}w^{\min}_{\tau^i-1}}
\leq \tfrac{(w^i)^{-\alpha}
(w^i)^{-\alpha\beta}}{(w^i)^{\frac{\alpha}{4}}w^i}
\eaqn
(note $w^{\min}_{\tau^i-1}\geq
(\tau^i-1)^{-\frac{1}{\alpha}}\geq
(w^i)^{(-\alpha)(-\frac{1}{\alpha})}$). Then
$(i)$ and
$(ii)$ follow from $\alpha=8$,
$\beta=0$, and
$(iii)$ and $(iv)$ follow from $\alpha=16$, $\beta=\frac{1}{16}$. The
asymptotic optimality finally follows from the Borel-Cantelli
Lemma, since according to $(ii)$ and $(iv)$,
\beqn
\Prob\bigg[
\frac{\loss\foe\leqT-\min_i\loss^i\leqT}{T}>
C T^{-\frac{1}{8}}\sqrt{\ln T}\,\bigg]\leq\frac{1}{T^2}
\eeqn
for an appropriate $C>0$.
\end{Proof}

As mentioned in the first paragraph of this section, it is
possible to avoid Lemma \ref{lemma:behbeh}, thus arriving
at better bounds. E.g.\ in $(i)$, choosing
$\tau^i=\big\lceil(\frac{1}{w^i})^8\big\rceil$,
$\gamma_t=t^{-\frac{1}{4}}$, and
$\eta_t=t^{-\frac{3}{4}}$, a regret bound of
$O\big((\frac{1}{w^i})^{11}+k^i T^{\frac{3}{4}}\big)$
can be shown. Of course, also a corresponding high
probability bound like $(ii)$ holds. Likewise, for a
similar statement as $(iii)$, we may set
$\tau^i=\big\lceil(\frac{1}{w^i})^{16}\big\rceil$,
$B_t=t^{\frac{1}{8}}$,
$\gamma_t=t^{-\frac{1}{4}}$, and
$\eta_t=t^{-\frac{3}{4}}$, arriving at a regret bound of
$O\big((\frac{1}{w^i})^{23}+k^i T^{\frac{3}{4}}\big)$
Generally, in this way any regret bound
$O\big((\frac{1}{w^i})^{c}+k^i T^{\frac{2}{3}+\eps}\big)$ is
possible, at the cost of increasing $c$ where $\eps\to 0$.

\section{Reactive environments and a universal master algorithm}\label{sec:active}

\emph{Regret} can become a quite subtle notion if we start
considering reactive environments, i.e.\ care for future
consequences of a decision. An extreme case is the ``heaven-hell"
example: We have two experts, one always playing 0 (``saying a
prayer"), the other one always playing 1 (``cursing"). If we
always follow the first expert, we stay in heaven and get no loss
in each step. As soon as we ``curse" only once, we get into hell
and receive maximum loss in all subsequent time steps. Clearly,
\emph{any} algorithm without prior knowledge must ``fail" in this
situation.

One way to get around this problem is taking into account the
\emph{actual} (realization of the) game we are playing. For
instance, after ``cursing" once, also the praying expert goes to
hell together with us and subsequently has maximum loss. Hence,
were are interested in a regret defined as $\Expect
\loss\leqT-\loss\leqT^i$ as in the previous section. So what is
missing? This becomes clear in the following example.

Consider the repeated ``prisoner's dilemma" against the
tit-for-tat\footnote{In the
prisoner's dilemma, two players both decide independently if
thy are \emph{cooperating (C)} or \emph{defecting (D)}. If
both play \emph{C}, they get both a small loss, if both play
\emph{D}, they get a large loss. However, if one plays \emph{C}
and one \emph{D}, the cooperating player gets a very large
loss and the defecting player no loss at all. Thus defecting
is a \emph{dominant} strategy. Tit-for-tat plays
\emph{C} in the first move and afterwards the opponent's
respective preceding move.} strategy \cite{Farias:03}. If we
use two strategies as experts, namely ``always cooperate" and
``always defect", then it is clear that always cooperating
will have the best long-term reward. However
standard expert advice or bandit master algorithm will not
discover this, since it compares only the losses in one step,
which are always lower for the defecting expert. To put it
differently, minimizing short-term regret is not at all a
good idea here. E.g. always defecting has no regret, while
for always cooperating the regret grows \emph{linearly}. But
this is only the case for short-term regret, i.e.\ if we
restrict to time intervals of length one.

\begin{figure}[t!]
\begin{center}
\fbox{
\begin{minipage}{\algowidth}
set $\tilde t=1$\\
For $t=1,2,3,\ldots$\\
\ii invoke $\FOE(t)$ and play its decision for $B_t$ basic time steps\\
\ii set $\tilde t=\tilde t+B_t$
\end{minipage}}
\caption{The algorithm $\FOEtilde$, where $B_t$ is specified in Corollary \ref{cor:active}.}
\label{fig:foetilde}
\end{center}
\end{figure}

We therefore give the control to a selected expert for
\emph{periods of increasing length}. Precisely, we introduce a new
time scale $\tilde t$ (the \emph{basic} time scale) at which we
have single games with losses $\tilde\loss_{\tilde t}$. The
master's time scale $t$ does not coincide with $\tilde t$.
Instead, at each $t$, the master gives control to the selected
expert $i$ for $B_t\geq 1$ single games and receives loss
$\loss_t^i=\sum_{\tilde t=\tilde t(t)}^{\tilde t(t)+B_t-1}
\tilde\loss_{\tilde t}^i$. (The points $\tilde t(t)$ in basic time
are defined recursively, see Fig. \ref{fig:foetilde}.) Assume that
the game has bounded instantaneous losses $\tilde\loss_{\tilde
t}^i\in[0,1]$. Then the master algorithm's instantaneous losses
are bounded by $B_t$. We denote the algorithm, which is completely
specified in Fig. \ref{fig:foetilde}, by $\FOEtilde$. Then the
following assertion is an easy consequence of the previous
results.

\begin{Cor}\label{cor:active}
Assume $\FOEtilde$ plays a repeated game with bounded
instantaneous losses $\tilde\loss_{\tilde t}^i\in[0,1]$. Choose
$\gamma_t=t^{-\frac{1}{4}}$, $\eta_t=t^{-\frac{3}{4}}$,
$B_t=\lfloor t^{\frac{1}{16}}\rfloor$ and
$\tau^i=\lceil(w^i)^{-16}\rceil$. Then for all experts $i$ and all
$\tilde T\geq 1$,
\bqan
\tilde \loss\foetilde_{1:\tilde T} & \leq & \tilde \loss_{1:\tilde T}^i+
  O\big((\tfrac{1}{w^i})^{22}+k^i\tilde
  T^{\frac{9}{10}}\big)
  \wprob 1-\tilde T^{-\frac{32}{17}} \und \\
  \Expect
  \tilde \loss\foetilde_{1:\tilde T} & \leq & \tilde
  \loss_{1:\tilde T}^i+O\big((\tfrac{1}{w^i})^{22}+k^i\tilde
  T^{\frac{9}{10}}\big).
\eqan
Consequently, $\limsup_{T\to\infty} (\tilde
\loss\foetilde_{1:\tilde T}-\tilde \loss^i_{1:\tilde T})/\tilde
T\leq0$ a.s. The rate of convergence is at least $\tilde
T^{-\frac{1}{10}}$, and it can be improved to $\tilde
T^{-\frac{1}{3}+\eps}$ at the cost of a larger power of
$\frac{1}{w^i}$.
\end{Cor}

\begin{Proof}
This follows from changing the time scale from $t$ to $\tilde t$
in Corollary \ref{cor:arbitrary}: $\tilde t$ is of order
$t^{1+\frac{1}{16}}$. Consequently, the regret bound is
$O\big((\tfrac{1}{w^i})^{22}+k^i\tilde T^{\frac{15}{17}}\sqrt{\ln
\tilde T}\big)\leq O\big((\tfrac{1}{w^i})^{22}+k^i\tilde
T^{\frac{9}{10}}\big)$.
\end{Proof}

Broadly spoken, this means that $\FOE_{\tilde T}$ performs
asymptotically as well as the best expert. Asymptotic performance
guarantees for the Strategic Experts Algorithm have been derived
in \cite{Farias:03}. Our results improve upon this by providing a
rate of convergence. One can give further corollaries, e.g.\ in
terms of flexibility as defined in \cite{Farias:03}.

Since we can handle countably infinite expert classes, we may
specify a \emph{universal} experts algorithm. To this aim, let
expert $i$ be derived from the $i$th (valid) program $p^i$ of some
fixed universal Turing machine. The $i$th program can be
well-defined, e.g.\ by representing programs as binary strings and
lexicographically ordering them \cite{Hutter:04uaibook}. Before
the expert is consulted, the relevant input is written to the
input tape of the corresponding program. If the program halts, an
appropriate part of the output is interpreted as the expert's
recommendation. E.g.\ if the decision is binary, then the first
bit suffices. (If the program does not halt, we may for
well-definedness just fill its output tape with zeros.) Each
expert is assigned a prior weight by
$w^i=2^{-\mbox{\scriptsize{length}}(p^i)}$, where
$\mbox{length}(p^i)$ is the length of the corresponding program
and we assume the program tape to be binary. This construction
parallels the definition of Solomonoff's \emph{universal prior}
\cite{Solomonoff:78}.

\begin{Cor}\label{cor:universal}
If $\FOEtilde$ is used together with a universal expert class as
specified above and the parameters $\eta_t,\gamma_t,B_t,\delta_T$
are chosen as in Corollary \ref{cor:active}, then it performs
asymptotically at least as well as \emph{any computable expert}
$i$. The upper bound on the rate of convergence is exponential in
the complexity $k^i$ and proportional to $\tilde
t^{-\frac{1}{10}}$ (improvable to $\tilde t^{-\frac{1}{3}+\eps}$).
\end{Cor}

The universal prior has been used to define a universal agent AIXI
in a quite different way \cite{Hutter:01aixi,Hutter:04uaibook}.
Note that like the universal prior and the AIXI agent, our
universal experts algorithm is not computable, since we cannot
check if a the computation of an expert halts. On the other hand,
if used with computable experts, the algorithm is computationally
feasible (at each time $t$ we need to consider only finitely many
experts). Moreover, it is easy to impose an additional constraint
on the computation time of each expert and abort the expert's
computation after $C_t$ operations on the Turing machine. We may
choose some (possibly rapidly) growing function $C_t$, e.g.\
$C_t=2^t$. The resulting master algorithm is fully computable and
has small regret with respect to all resource bounded strategies.

It is important to keep in mind that Corollaries \ref{cor:active}
and \ref{cor:universal} give assertions relative to the experts'
performance merely on the \emph{actual} action-observation
sequence. In other words, if we wish to assess how well
$\FOEtilde$ does, we have to evaluate the actual \emph{value} of
the best expert \cite{Farias:04}. Note that the whole point of our
increasing time construction is to cause this actual value to
coincide with the value under \emph{ideal} conditions. For passive
tasks, this coincidence always holds with any experts algorithm.
With $\FOEtilde$, the actual and the ideal value of an expert
coincide in many further situations, such as ``finitely
controllable tasks". By this we mean cases where the best expert
can drive the environment into some optimal state in a fixed
finite number of time steps. An instance is the prisoner's dilemma
with tit-for-tat being the opponent. The following is an example
for a formalization of this statement.

\begin{Prop}
Suppose $\FOEtilde$ acts in a (fully or partially observable)
Markov Decision Process. Let there be a computable strategy which
is able to reach an ideal (that is optimal w.r.t.\ reward) state
sequence in a fixed number of time steps. Then $\FOEtilde$
performs asymptotically optimal.
\end{Prop}

This statement may be generalized to cases where only a close to
optimal state sequence is reached with high probability. However,
we need assumptions on the closeness to optimality for a given
target probability, which are compatible with the sampling
behavior of $\FOEtilde$.

Not all environments have this or similar nice properties. As
mentioned above, any version of $\FOE$ would not perform well in
the ``heaven-hell" example. The following is a slightly more
interesting variant of the heaven-hell task, where we might wish
to learn optimal behavior, however $\FOE$ will not. Consider the
heaven-hell example from the beginning of this section, but assume
that if at time $t$ I am in hell and I ``pray" for $t$ consecutive
time steps, I will get back into heaven. Then it is not hard to
see that $\FOE$'s exploration is so dominant that almost surely,
$\FOE$ will eventually stay in hell.

Simulations with some $2\times 2$ matrix games show similar
effects, depending on the opponent. We briefly discuss the
repeated game of ``chicken"\footnote{ This game, also known as
``Hawk and Dove", can be interpreted as follows. Two coauthors
write a paper, but each tries to spend as little effort as
possible. If one succeeds to let the other do the whole work, he
has a high reward. On the other hand, if no one does anything,
there will be no paper and thus no reward. Finally, if both decide
to cooperate, both get some reward. We choose the loss matrix as
${1\quad 0 \choose 0.8\ 0.5}$, the learner is the column player,
the opponent's loss matrix is the transpose, choosing the fist
column means to defect, the second to cooperate. Hence, in the
repeated game, it is socially optimal to take turns cooperating
and defecting.}. In this game, it is desirable for the learner to
become the ``dominant defector", i.e.\ to defect in the majority
of the cases while the opponent cooperates. Let's call an opponent
``primitive" if he agrees to cooperate after a fixed number of
consecutive defecting moves of $\FOE$, and let's call him
``stubborn" if this number is high. Then $\FOE$ learns to be the
dominant defector against any primitive opponent, however
stubborn. On the other hand, if the opponent is some learning
strategy which also tries to be the dominant defector and learns
faster (we conducted the experiment with AIXI
\cite{Hutter:04uaibook}), then $\FOE$ settles for cooperating, and
the opponent will be the dominant defector. Interestingly however,
AIXI would not learn to defect against a stubborn primitive
opponent. Under this point of view, it seems questionable that
there is something like a universally optimal balance of
exploration vs.\ exploitation in active learning at all.

\section{Discussion}\label{sec:discussion}

\absatz{An alternative argument for adaptive adversary.} As
mentioned in the beginning of Section \ref{sec:master}, the
analysis we gave uses a trick from \cite{McMahan:04}. Such a trick
seems necessary, as the basic FPL analysis only works for
oblivious adversary. The simple argument from
\cite{Hutter:05expertx} which we used in the last paragraph of the
proof of Lemma \ref{lemma:ifplbeh} works only for full observation
games (note that considering the estimated losses, we were
actually dealing with full observations there). In order to obtain
a similar result in the partial observation case, we may argue as
follows. We let the game proceed for $T$ time steps with
independent randomization against an adaptive adversary. Then we
analyze $\FOE$'s performance \emph{in retrospective}. In
particular, we note that for the losses assigned by the adversary,
$\FOE$'s expected regret coincides with the regret of another,
virtual algorithm, which uses (in its FPL subroutine) identical
perturbations $q_t\equiv q$. Performing the analysis for this
virtual algorithm, we arrive at the desired assertion, however
without needing Lemma \ref{lemma:behbeh}. This results in tighter
bounds as stated above. The argument is formally elaborated in
\cite{Poland:05fpla}.

\absatz{Actual learning speed and lower bounds.} In practice, the
bounds we have proven seem irrelevant except for small expert
classes, although asserting almost sure optimality and even a
convergence rate. The exponential of the complexity
$\frac{1}{w^i}$ may be huge. Imagine for instance a moderately
complex task and some good strategy, which can be coded with mere
500 bits. Then its prior weight is $2^{-500}$, a constant which is
not distinguishable from zero in all practical situations. Thus,
it seems that the bounds can be relevant at most for small expert
classes with uniform prior. This is a general shortcoming of
bandit style experts algorithms: For uniform prior a lower bound
on the expected loss which scales with $\sqrt n$ (where $n$ is the
size of the expert class) has been proven \cite{Auer:02bandit}.

In order to get a lower bound on $\FOE$'s regret in the time $T$,
observe that $\FOE$ is a \emph{label-efficient} learner
\cite{Cesa:04,Cesa:04partial}: According to the definition in
\cite{Cesa:04partial}, we may assume that in each exploration
step, we incur maximal loss $B_t$. It is immediate that the same
analysis then still holds. For label-efficient prediction,
Cesa-Bianchi et al. \cite{Cesa:04partial} have shown a lower
regret bound of $O(T^{\frac{2}{3}})$. Since according to the
remark at the end of Section \ref{sec:master}, we have an upper
bound of $O\big((\frac{1}{w^i})^{c}+k^i
T^{\frac{2}{3}+\eps}\big)$, this is almost tight except for the
additive $(\frac{1}{w^i})^{c}$ term. It is an open problem to
state a lower bound simultaneously tight in both $\frac{1}{w^i}$
and $T$.

Even if the bounds, in particular $\frac{1}{w^i}$, seem not
practical, maybe $\FOE$ would learn sufficiently quickly in
practice anyway? We believe that this is not so in most cases: The
design of $\FOE$ is too much tailored towards worst-case
environments, $\FOE$ is too \emph{defensive}. Assume that we have
a ``good" and a ``bad" expert, and $\FOE$ learns this fact after
some time. Then it still would spend a relatively huge fraction of
$\gamma_t=t^{-\frac{1}{4}}$ to exploring the bad expert. Such
defensive behavior seems only acceptable if we are already
starting with a class of good experts.

\absatz{Acknowledgment.}
This work was supported by SNF grant 2100-67712.02
and JSPS 21st century COE program C01.


\end{document}